\documentclass[runningheads]{llncs}
\usepackage{amsmath,amssymb,amsfonts}

\usepackage{amsthm}
\usepackage{mathtools}

\usepackage{graphicx}
\usepackage{array}
\usepackage{multirow}
\usepackage{booktabs}
\usepackage{algorithm}
\usepackage{algpseudocode}
\usepackage[export]{adjustbox}

\usepackage{cite}

\usepackage{hyperref}
\hypersetup{
  hidelinks,
  breaklinks=true,
  colorlinks=true,
  urlcolor=blue,
  bookmarksopen=false,
  pdftitle={Title},
  pdfauthor={Author}
}
\usepackage{cleveref}
\usepackage{wrapfig}

\usepackage[font=small,skip=10pt]{caption}

\crefname{algorithm}{Alg.}{Algs.}
\crefname{figure}{Fig.}{Figs.}
\crefname{section}{Sec.}{Secs.}
\crefname{theorem}{Thm.}{Thms.}
\crefname{proposition}{Prop.}{Props.}
\crefname{table}{Tab.}{Tabs.}
\crefname{appendix}{App.}{Apps.}

\makeatletter
\def\@fnsymbol#1{\ensuremath{\ifcase#1\or \bigstar \or \dagger \or \ddagger \else\@ctrerr\fi}}
\makeatother

\newcommand{\image}{\textit{m}}
\newcommand{\model}{\textit{f}}
\newcommand{\IoU}{\text{IoU}}

\newcommand{\mysubsection}[1]{%
    \medskip
    \noindent
    \textbf{#1}
}

\begin{document}

\title{Towards Formal Verification of \texorpdfstring{\\}{ } Deep Neural Networks for Object Detection}

\author{
    Avraham Raviv\inst{1}\thanks{These authors contributed equally.} \and
    Yizhak Y. Elboher\inst{2}\textsuperscript{$\bigstar$}\thanks{Corresponding author.} 
    \and     
    Omri Isac\inst{2,3} \and
    Michelle Aluf-Medina\inst{1} \and
    Ben Hagag\inst{3}\and
    Golan Shmueli\inst{3}\and
    Guy Katz \inst{2,3} \and
    Hillel Kugler\inst{1}
}

\authorrunning{A. Raviv, YY. Elboher et al.}
\institute{
    Bar-Ilan University, \email{\{ravivav1, alufmem, hillelk\}@biu.ac.il} \and
    The Hebrew University of Jerusalem, \email{\{yizhak.elboher,omri.isac,g.katz\}.mail.huji.ac.il} \and
    Rigora.ai, \email{Bhagag@andrew.cmu.edu, golan.shmueli@rigora.ai}
}

\titlerunning{Formal Verification of OD}

\maketitle
\def\thefootnote{$\bigstar$}
\def\thefootnote{\arabic{footnote}} 

\begin{abstract}
Deep neural networks (DNNs) are widely used in real-world computer vision
applications, yet they remain vulnerable to errors and adversarial
attacks.  Formal verification offers a systematic approach to
identify and mitigate these vulnerabilities, enhancing model
robustness and reliability.  While most existing verification
methods focus on image classification models, this work extends
formal verification to the more complex domain of \emph{object
detection} models. We propose a formulation for verifying the
robustness of such models and demonstrate how state-of-the-art
verification tools, originally developed for classification, can be
adapted for this purpose. Through a comprehensive evaluation,
we highlight the ability of formal verification
to uncover vulnerabilities in object detection models, and derive formal robustness guarantees, underscoring
the potential and need to further extend verification efforts in this domain. This work
lays the foundation for further research into formal verification of
object detection models across a broader range of computer vision
applications. Our source code is publicly available online.\footnote{\label{repo}\url{https://github.com/AvrahamRaviv/FVOD_2025}}
\end{abstract}


\section{Introduction}
Deep neural networks (DNNs) have demonstrated exceptional
capabilities~\cite{KrSuHi12}, consistently achieving state-of-the-art
performance across diverse domains, including computer vision
(CV)~\cite{KrSuHi12,DoBeKoWeZhUnDeMiHeGeUsHo21}, natural language
processing
(NLP)~\cite{VaShPaUsJoGoKaPo23,DeChLeTo19,BrMaRySuKaDhNeShSaAsAgHeKrHeChRaZiWuWi20},
audio~\cite{GuQiChPaZhYuHaWaZhWuPa20,BaZhMoAu20}, and video
analysis~\cite{LiNiCaWeZhLiHu21,ArDeHeSuLuSc21}. These achievements
underscore the transformative potential of DNNs, enabling
groundbreaking advancements in technology and applications. However,
despite their success, the deployment of DNNs in safety-critical
industries such as autonomous vehicles, aviation, financial services,
and healthcare remains limited. One major hindrance to the deployment of DNNs is their
inherently statistical nature and dependence on training data and methods, which makes them vulnerable to both benign and adversarial input perturbations. Even minor, often
imperceptible, alterations to a DNN's input can result in significant errors, posing substantial risks in safety-critical applications.

Object detection (OD) plays a central role in many safety-critical 
applications, from autonomous driving to medical imaging. There, a DNN is required both to classify objects and to localize them within boxes. 
Compared to classification, OD introduces additional challenges: 
predictions must satisfy both spatial accuracy (e.g., measured by 
domain-specific thresholds) and label correctness, while being robust 
to spurious detections. These properties make OD particularly 
vulnerable to adversarial perturbations, underscoring the need for 
rigorous verification techniques towards developing safe and robust systems. 

The vulnerability of DNNs has catalyzed the creation of an entire branch of research dedicated to understanding and mitigating these weaknesses through both attack and defense mechanisms. Among these, formal verification of DNNs has emerged as a promising approach~\cite{KaBaDiJu17,LiArLaStBaKo20}. Formal methods aim to mathematically prove the robustness of DNNs against adversarial perturbations, providing stronger guarantees that cannot be achieved by empirical testing. Initially, the focus of DNN verification was largely confined to classification networks~\cite{KaBaDiJu17,ZhWeChHsDa18}. Over time, the scope has expanded to encompass additional areas such as NLP~\cite{CaDiKoArIsDaKaRiLem24}, reinforcement learning~\cite{AmScKa21,BaAmCoReKa23}, and explainability~\cite{BaKa23}. Nevertheless, the application of formal verification techniques to other crucial CV tasks, particularly object detection and segmentation, remains underexplored despite their importance in numerous practical applications.

\mysubsection{Our Contributions.}
This work systematically studies the robustness of DNNs in object detection, towards bridging a key gap in formal verification. Our main contributions are:
\begin{enumerate}
    \item We formally define robustness properties for object detection and cast them as verification problems, providing a structured framework for analyzing model vulnerabilities.
    \item We present a first sound and complete method for robustness queries of object detection models.
    \item We extend state-of-the-art verification tools, originally built for classification, to handle these queries with minimal modifications. For transparency and reproducibility, we make our code available online.\footref{repo}
    \item We conduct extensive experiments demonstrating (i) adversarial vulnerabilities; (ii) comparative robustness across models and datasets; (iii) superior scalability to interval bound propagation (IBP) based methods in the number of solved queries; and (iv) robustness verification of a model detecting multiple objects.
\end{enumerate}

\begin{figure}[ht!]
  \centering
  \includegraphics[width=0.7\linewidth, keepaspectratio]{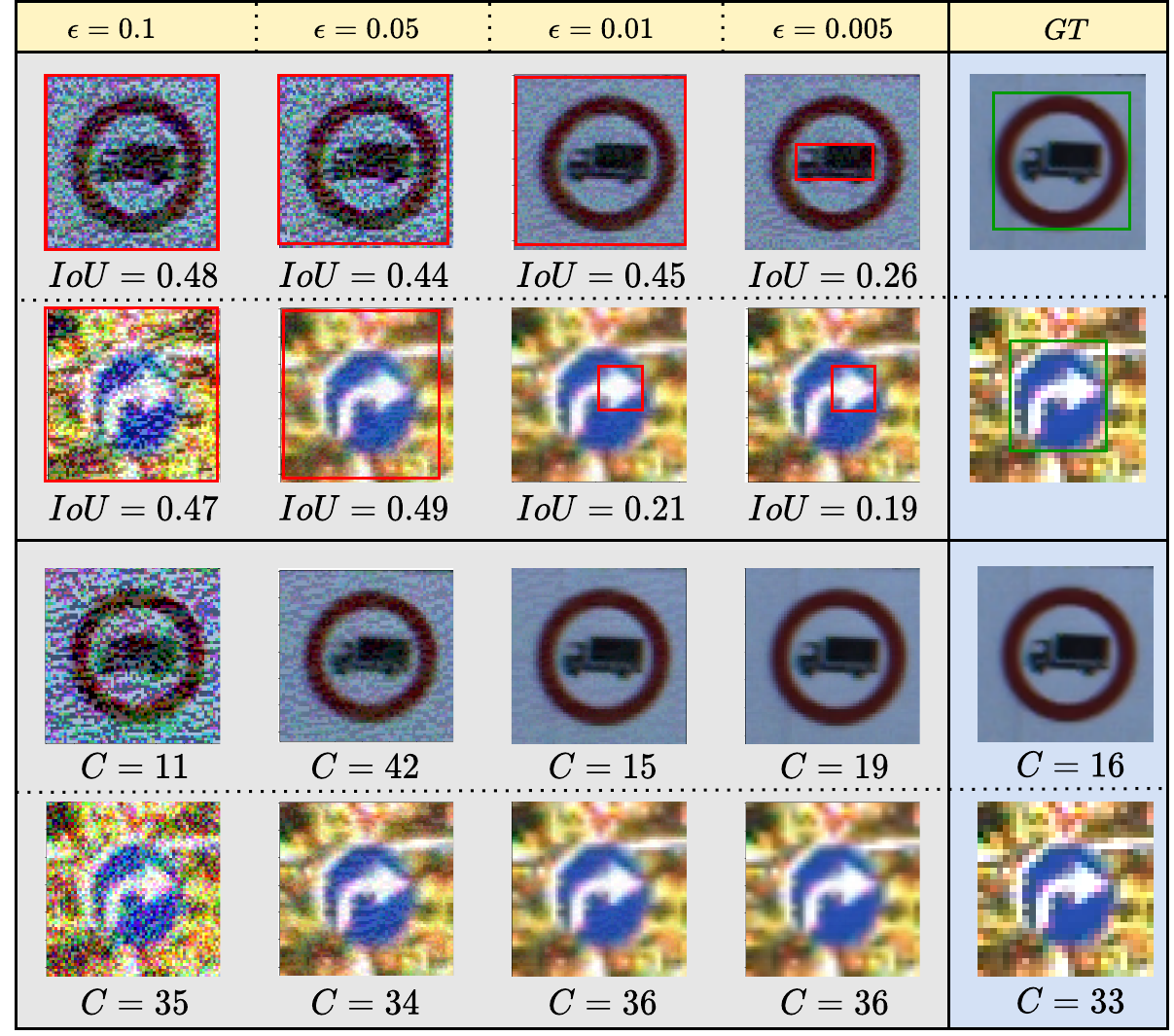}
  \caption{Examples of attacked images. The rightmost column shows original images, while the others display attacked images with different noise levels (maximum allowed noise $\epsilon$). The top two rows illustrate misdetection,
  where the object’s location was not detected sufficiently close to the ground truth as measured by the intersection over union (IOU),
  and the bottom two rows show misclassification into class $C$ (class mapping is detailed in \cite{WeKa17}).}
  \label{Figure_intro}
\end{figure}

\cref{Figure_intro} illustrates some of the robustness threats
identified by our method. These examples highlight the critical
challenges associated with verifying detectors and the
potential impact of our approach in addressing these challenges.

The remainder of this paper is structured as follows. \cref{sec:RW} reviews related work on neural network verification and adversarial attacks in CV. ~\cref{sec:prelim} provides background on object detection and the formal verification of DNNs. ~\cref{sec:OD_as_FV_problem} details our methodology, including the formulation of the object detection task as a formal verification problem (\cref{encoding}), and the adaptation of verification tools for single-object (\cref{single_OD}) and multi-object (\cref{multi_OD}) detection. ~\cref{sec:evaluation} presents our evaluation results, demonstrating the effectiveness of our approach across multiple models and datasets. Finally, 
~\cref{sec:Summary} concludes the paper with a summary of our contributions and potential directions for future research.

\section{Related Work}
\label{sec:RW}
This work aims to address the gap between object detection and the formal verification of neural networks, bridging two critical areas for machine learning research. State-of-the-art object detection algorithms, including Faster R-CNN~\cite{ShKaRoJi15}, YOLO~\cite{ReDiGiFa16}, and SSD~\cite{LiAnErSzReFuBe15}, rely on deep convolutional networks to achieve impressive accuracy and computational efficiency across diverse datasets. Despite their effectiveness, these models remain vulnerable to adversarial threats. Beyond common classification attacks~\cite{KuGoBe17}, object detection systems are susceptible to patch-based attacks~\cite{RoAbGi17,ShBiMuNa22}, where adversarial patches are introduced into the image, and adversarial camouflage~\cite{HuGaZhXiYuZoLi20,ZhFoDaGo18}, which subtly modifies object appearances to render them ``invisible'' to the detector. These vulnerabilities underscore the importance and relevance of our investigation.

Our research builds upon the extensive body of work on the formal verification of deep neural networks. Various verification approaches and tools have been developed, utilizing techniques such as SMT solving~\cite{KaHuIbJuLaLiShThWuZeDiKoBa19, WuIsZeTaIsZeTaDaKoReAmJuBa24}, symbolic interval propagation~\cite{HeLo20,HeLo21}, and optimization-based methods like MILP and LP solving~\cite{DuJhSaTi17,TjXiTe17,FiJo18,BuTuToKoMu18,LoMa17}. Abstract interpretation has also been used, with methods incorporating symbolic analysis~\cite{ZhWeChHsDa18}, interval propagation~\cite{WaZhXuLiJaHsKo21}, and advanced abstraction techniques~\cite{FeNiJoVe22,MuMaSiPuVe22}. To enhance computational efficiency and scalability, many of these approaches integrate branch-and-bound mechanisms~\cite{LaDo1960,MoJaSaSe16,BuLuTuToKoKu20}. In our experiments, we utilize the Alpha-Beta-CROWN verifier, a state-of-the-art tool that supports networks with multiple outputs and various activation functions.

The scope of formal verification for DNNs extends beyond computer vision  and has found applications in other domains, including natural language processing (NLP)~\cite{CaDiKoArIsDaKaRiLem24}, reinforcement learning~\cite{AmCoYeMaHaFaKa23,YeAmElHaKaMa22,KaBaKaSc19,RaBrReAlKu23,RaGeBeAlKu23}, and explainability~\cite{BaKa23,BaAmCoReKa23}. Additional areas of application include generalization~\cite{AmMaZeKaSc23,AmMaZeKaSc24}, delta-debugging~\cite{ElKa23}, and robustness assessment~\cite{ElElIsDuGaPoBoChKa24}. Formal verification techniques have also been extended in multiple directions, such as proof production~\cite{IsBaZhKa22}, abstraction refinement~\cite{YeAmElHaKaMa23,RaBrReAlKu23,ElGoKa20,CoElBaKa23,ElIsKaLaWu25,BaElLaAlKa25}, and residual reasoning~\cite{ElCoKa22}. These advancements highlight the complementary nature of different verification approaches, and exploring potential synergies between them and our methodology is a promising avenue for future research.

Among the works most closely related to ours is the study by Cohen et al.~\cite{CoDuBoGaPaPuGa24}, which investigates the formal verification of DNN-based object detection systems using abstract domains for IoU constraints. Their findings highlight the practical importance of verification in safety-critical settings, as evidenced by industrial partners such as Airbus. Chiang et al.~\cite{Chiang2020Detection} propose a complementary approach, \emph{Detection as Regression}, which achieves certified guarantees via median smoothing, but remains sound and incomplete. Other recent efforts have explored certified object detection in applied domains, including conformal prediction for runway detection~\cite{Zouzou2025ConformalRunway}, certified YOLO-based runway detection~\cite{Cheng2024CertifiedRunwayYOLO}, and runtime monitoring for out-of-distribution detection in such networks~\cite{Hashemi2023Runtime}. However, none of these methods is complete.

\section{Preliminaries}\label{sec:prelim}

In computer vision, a variety of tasks require the analysis and interpretation of visual data, each presenting its own unique challenges. Among the foundational tasks is \emph{image classification}, where models categorize entire images into predefined classes. Formally, a model $\model$ takes an input image $\image_0\in\mathbb{R}^d$ and maps it to a class label $c$ selected from a discrete set of possible labels $\mathcal{C}$: $\model(\image_0)\in \mathcal{C}$. Each label is associated with a confidence score, and the highest score determines the predicted class.

Building on classification, \emph{object detection} (OD) extends the task beyond identifying the classes present in an image, to also include localizing each object within a predicted bounding box (bbox). An OD model processes an input image $\image_0$ to produce a set of bboxes $\mathcal{B}$. Each bbox in $\mathcal{B}$ is characterized by a tuple ${(x, y, h, w, c)}$, where ${(x, y)}$ represents the coordinates of the top-left corner, ${(h, w)}$ denotes its dimensions, height and width, respectively, and $c$ specifies the detected class label.

Since perfect matching between predictions and ground truth is often unattainable, the comparison is assessed using the \emph{intersection over union} (\IoU{}), a metric defined as the ratio between the intersection and union of respective bboxes. If the \IoU{} surpasses a predefined threshold $\tau$ and the class is correct, the prediction is deemed accurate.

An \emph{adversarial attack} intentionally alters an original data point $\image_0 \in \mathbb{R}^d$ into a perturbed version $\image \in \mathbb{R}^d$, where $\image \neq \image_0$. Such alterations are designed to mislead the model, causing it to produce incorrect or unintended outputs.

In \textit{formal verification}, the objective is to ascertain that a neural network adheres to specified properties for all possible inputs within a defined set. This input set $\mathcal{I}$ is typically bounded by constraints $\mathcal{P}$ that restrict the allowed input variations. For instance, for an input $\mathbf{x} \in \mathbb{R}^d$, $\mathcal{I}$ might encompass all points $\mathbf{x} \in \mathbb{R}^d$ such that the deviation from a base input $\mathbf{x}_0$ is limited by $|\mathbf{x} - \mathbf{x}_0| \leq \delta$, where $\delta$ denotes the permissible perturbation magnitude. The model's output $\mathcal{O} = {\model(\mathbf{x}):{\mathbf{x} \in \mathcal{I}}}$ is then evaluated against output constraints $\mathcal{Q}^{neg}$, which specify undesired behaviors. 
A \textit{verification query} is, therefore, a triplet $(\model, \mathcal{P}, \mathcal{Q}^{neg})$. Verification ensures that no inputs within $\mathcal{I}$ that satisfy $\mathcal{P}$ produce outputs satisfying $\mathcal{Q}^{neg}$. Formally:
\[ \lnot\exists \image \in \mathcal{P}, \quad \model(\image) \in \mathcal{Q}^{neg} \]
This guarantees the network behaves correctly for all permissible inputs, thus providing robust reliability assurances.

In the context of image classification, verification involves ensuring consistent predictions under input variations. The constraints $\mathcal{P}$ typically define allowed modifications to an image $\mathbf{x}_0$, forming a set $\mathcal{I}$ of perturbed inputs, while $\mathcal{Q}=\lnot\mathcal{Q}^{neg}$ mandates consistent class predictions across $\mathcal{I}$. Verification in CV has largely focused on classification tasks, emphasizing robustness against permissible perturbations.

\mysubsection{Object Detection: Threat Model.}
In this work, we examine a threat model where an attacker can perturb the input by introducing small alterations to an image, with access to the model's parameters (white-box access) and an unlimited number of queries. Unlike classification, which typically revolves around a single objective, misclassification, object detection presents a more diverse set of attack vectors, often grouped into three categories 
(e.g., \cite{Miller2022ODFail}), including:

\begin{enumerate}
\item \emph{Misdetection} (bbox false negatives): This involves manipulating the model to fail in detecting genuine objects within the image.
\item \emph{Misclassification} (classification error): This mirrors adversarial attacks in classification, where the aim is to mislabel detected objects or miss them entirely. Notably, misdetection is a special case of misclassification.
\item \emph{Overdetection} (bbox false positives): Here, the objective is to deceive the model into detecting nonexistent objects within the image.
\end{enumerate}

\section{Object Detection Robustness as a Formal Verification Problem}\label{sec:OD_as_FV_problem}
To express robustness against the aforementioned attacks as a formal verification problem, we define suitable input constraints (\(P\)) and output constraints (\(Q\)) that characterize when an attack is impossible. The verification objective is to prove the implication \(P \Rightarrow Q\). If the input satisfies \(P\), then the output must satisfy \(Q\). If this implication fails, the verifier returns a counterexample, demonstrating a feasible attack and providing an adversarial example.

We define each adversarial attack vector in our threat model as a logical \(P \Rightarrow Q\) condition. In~\cref{encoding}, we show how these abstract definitions are translated into concrete verification queries via input and output constraints compatible with formal verification tools.

We first consider single-object detection in~\cref{single_OD}, where the goal is to detect a single object and accurately predict its bbox. We then extend the approach to multi-object detection in~\cref{multi_OD}. Interestingly, in the single-object setting, overdetection and misclassification are equivalent, a result that contrasts with the intuitive expectation that overdetection aligns more closely with misdetection. We formalize each attack type to clarify this behavior.
We denote by $o_g$ the number of ground-truth objects and by $o_d$ the number of detected objects; in the single-object case, $o_g = o_d = 1$.

\mysubsection{Formal Definition of Object Detection Attacks.}
\label{subsec:Attacks Formal Definition}
For all the types of attacks considered in this work, the underlying threat model assumes that only a small perturbation can be applied to the original input image. Consequently, the input property must ensure that the perturbed input remains within a small distance from the original input. Formally, this input constraint is defined as: 
\[
P := |x - x_0| < \epsilon,
\] 
where \(\epsilon\) represents the maximum allowed perturbation magnitude, and \(x_0\) is the original input image. This ensures that the adversarial image \(x\) remains sufficiently close to the original image \(x_0\), preserving its semantic similarity.

For a given input image,  assume that there exist \(o_g\) ground truth objects represented by the set of tuples \(\{G_1, G_2, \dots, G_{o_g}\}\). Each ground truth object \(G_i\) is defined as a tuple \((x^g_i, y^g_i, h^g_i, w^g_i, c^g_i)\), where \(x^g_i\) and \(y^g_i\) specify the location of the bbox, \(h^g_i\) and \(w^g_i\) represent its height and width, and \(c^g_i\) is the associated class label. Similarly, the output of the object detection model is represented by \(o_d\) detected objects given as a set of tuples \(\{D_1, D_2, \ldots, D_{o_d}\}\), where each detected object \(D_j\) is defined as \((x^d_j, y^d_j, h^d_j, w^d_j, c^d_j)\), containing the bbox location, dimensions, and predicted class.

We now provide formal definitions for each of the three attack vectors: 

\begin{enumerate}
\item \textbf{Misdetection}: This attack occurs when at least one ground truth object \(G_i\) is not detected by the model. This happens when there is no detected bbox \(D_j\) that sufficiently overlaps with \(G_i\) (as determined by the IoU threshold \(\tau\)). Formally, the attack condition is defined as:
\small
\begin{gather*}
Q_1 := (o_g > o_d) \lor \bigg(\exists i \in [o_g] \ \forall j \in [o_d]:\ \IoU{}(G_i, D_j) < \tau \bigg).
\end{gather*}
\normalsize
To ensure robustness against misdetection, the model must satisfy \(P \Rightarrow \lnot Q_1\), which negates $Q_1$:
\small
\begin{gather*}
\lnot Q_1 := (o_g \leq o_d) \land \bigg(\forall i \in [o_g] \ \exists j \in [o_d]:\ \IoU{}(G_i, D_j) > \tau \bigg).
\end{gather*}
\normalsize

\item \textbf{Misclassification}: This attack occurs when a ground truth object is either not detected or is detected with an incorrect class label. Formally, it can be described as a combination of detection failure (bbox not detected) and classification failure (bbox detected but with an incorrect class). The condition for this attack is given by:
\small
\begin{gather*}
Q_2 := (o_g > o_d) \lor \bigg(\exists i \in [o_g] \ \forall j \in [o_d]:\ (\IoU{}(G_i, D_j) < \tau) \lor (c^g_i \neq c^d_j) \bigg).
\end{gather*}
\normalsize

The robustness of the model against misclassification is ensured by satisfying \(P \Rightarrow \lnot Q_2\), which negates $Q_2$:
\small
\begin{gather*}
\lnot Q_2 := (o_g \leq o_d) \land \bigg(\forall i \in [o_g] \ \exists j \in [o_d]:\ (\IoU{}(G_i, D_j) > \tau) \land (c^g_i = c^d_j) \bigg).
\end{gather*}
\normalsize

\item \textbf{Overdetection}: This type of attack arises when the model detects additional objects that are not present in the ground truth. The attack is characterized by the presence of detected bboxes that do not correspond to any ground truth objects. Formally, the attack condition is:
\small
\begin{gather*}
Q_3 := (o_d > o_g) \lor \bigg(\exists j \in [o_d] \ \forall i \in [o_g]:\ (\IoU{}(D_j, G_i) < \tau) \lor (c^g_i \neq c^d_j) \bigg).
\end{gather*}
\normalsize
To guarantee robustness against overdetection, the model must satisfy \(P \Rightarrow \lnot Q_3\), which negates $Q_3$:
\small
\begin{gather*}
\lnot Q_3 := (o_d \leq o_g) \land \bigg(\forall j \in [o_d] \ \exists i \in [o_g]:\ (\IoU{}(D_j, G_i) > \tau) \land (c^g_i = c^d_j) \bigg).
\end{gather*}
\normalsize
\end{enumerate}

The robustness properties \(P \Rightarrow \lnot Q_1\), \(P \Rightarrow \lnot Q_2\), and \(P \Rightarrow \lnot Q_3\) respectively ensure the model's robustness against misdetection, misclassification, and overdetection attacks. However, it is important to note that these properties rely on constraints involving the IoU, which is not inherently calculated by the neural network model. Therefore, for the verification query to be well-defined and executable, the output constraints must be reformulated to deal exclusively with values that correspond to existing neurons within the network.

\mysubsection{Generating Equivalent Verification Queries.}
\label{subsec:Equivalent Verification Query}
To create a standard and well-defined verification query corresponding
to each attack type, we modify both the model architecture and the
output constraints. Specifically, we transform a query of the form
\((\model, P, \lnot Q_i)\)
into an equivalent query \((f_i', P, \lnot Q_i')\), where \(f_i'\) is an extended version of the original network. We add neurons to encode the predicate $\mathrm{IoU}>\tau$ via $A_I - \tau\cdot A_U > 0$; the full property then combines this predicate with the relevant class constraints. By construction, this ensures that the updated output constraint \(\lnot Q_i'\) in \(f_i'\) aligns with the semantics of \(\lnot Q_i\) in the original model \(\model\). 

The reformulation allows us to work within the framework of modern verification tools, avoiding direct computation of complex operations, such as those involving the intersection-over-union (\IoU{}) metric, within the neural network. Instead, the equivalent condition is encoded in a way that modern verifiers can efficiently handle. 

\subsection{Intersection and Union Encoding}
\label{encoding}

\mysubsection{Modeling \IoU{}-related Conditions with Neural Layers.}
A key component of our approach is extending the given network with additional layers that encode a condition equivalent to \(\IoU{} > \tau\). Directly computing \IoU{} within a single neuron is non-trivial, as it involves the ratio between the intersection and union areas of two bboxes, both unknown a priori, and requires division, an operation unsupported by most neural network verification frameworks.

To overcome this limitation, we reformulate the condition to avoid explicit division. The equivalent constraint is expressed using only subtraction and multiplication by known scalars, which can be naturally represented using neural layers and are fully supported by modern verification tools.

The reformulated condition is encoded by appending additional layers after the original output layer, while leaving the original network architecture unchanged. These layers compute two quantities: the intersection area \(A_I\) and the union area \(A_U\), each realized by a dedicated neuron.

Finally, an additional layer evaluates the equivalent \(\IoU{} > \tau\) condition, producing a positive output if and only if the condition holds. This construction enables verification of \IoU{}-based properties without division, relying solely on verifier-compatible operations.

\mysubsection{Intersection and Union.}
Given a pair of ground truth bbox and predicted bbox, \( B_{gt} = (x_g, y_g, h_g, w_g, c_g) \) and \( B_{dt} = (x_d, y_d, h_d, w_d, c_d) \), the calculation of intersection and union values involves only \textit{max}, \textit{min}, \textit{add} and \textit{subtract} operations. 
The intersection between \( B_{gt} \) and \( B_{dt} \) is a rectangle $R$ with the following coordinates:
\begin{align*}
x_{i1} &= \max(x_g, x_d) & x_{i2} &= \min(x_g + w_g, x_d + w_d)\\
\ y_{i1} &= \max(y_g, y_d) & y_{i2} &= \min(y_g + h_g, y_d + h_d)
\end{align*}
The width (\( w_{i} \)) and height (\( h_{i} \)) of $R$, and the implied area of the intersection, denoted by \(A_{I}\), are given by:
\begin{align*}
w_{i} &= \max(0, x_{i2} - x_{i1}) \hspace{0.5em} h_{i} = \max(0, y_{i2} - y_{i1}) \hspace{0.5em} A_{I}=w_i\cdot h_i
\end{align*}
The areas of the ground truth bbox (\( A_{g} \)) and the predicted bbox (\( A_{d} \)), as well as the implied area of the union (\( A_{U} \)) are given by:
\begin{align*}
A_g &= w_g \cdot h_g & A_d &= w_d \cdot h_d & A_{U} = A_g + A_d - A_{I}
\end{align*}

\mysubsection{\IoU{} and Equivalent Constraint.}
The IoU is defined as:
\[
\IoU{} = \frac{A_{I}}{A_{U}}
\]
We should handle the \IoU{} constraint which appears in misdetection, misclassification and overdetection: $\IoU{}>\tau$. We represent this constraint with an equivalent linear constraint, which can be easily encoded with neural layers:

\begin{align*}
\IoU{}>\tau\ & \Leftrightarrow \frac{A_{I}}{A_{U}} > \tau \Leftrightarrow \ A_{I} > \tau \cdot A_{U} \Leftrightarrow A_{I} - \tau \cdot A_{U} > 0
\end{align*}

\begin{figure}[ht!]
    \centering
    \includegraphics[width=0.6\columnwidth]{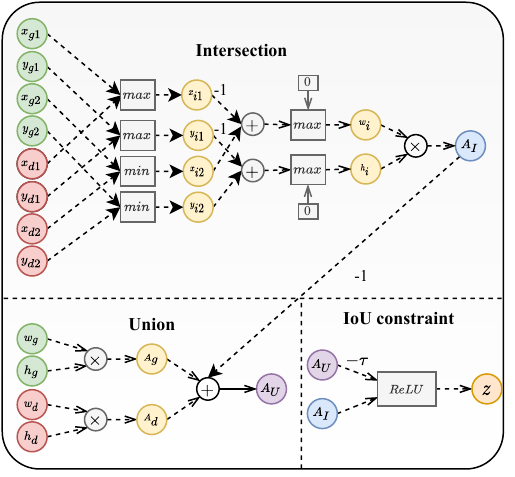}
    \vspace{-0.2cm}
  \caption{Encoding the \IoU{} operation. At first, the area of the
    intersection and the area of the union are extracted into a couple of neurons. Then, the equivalent condition is encoded using a single neural layer, and the output $z$ is positive if and only if $\IoU{}>\tau$.}
      \vspace{-0.2cm}
\label{fig:encoding_equivalent_iou_condition}
\end{figure}

\cref{fig:encoding_equivalent_iou_condition} illustrates how the \IoU{} between a single ground truth bbox and a single predicted bbox can be encoded with neural layers. From now on, we denote with $z$ the variable whose result is positive if and only if the condition $\IoU{}>\tau$ holds.

\subsection{Single Object detection}
\label{single_OD}
We now return to the attacks discussed above and show their full encoding as verification problems for the single-object case.
\subsubsection{Misdetection.} In this case, the
output condition reduces to $\IoU{}>\tau$. In order to adapt the output
constraint to the new network with equivalent constraint, we
define $z>0$ as the new output constraint, yielding the query $(f',P,z>0)$.

\subsubsection{Misclassification.}
In this case, there are two conditions involved: $(i)$ $\IoU{}>\tau$ and $(ii)$ $c_g = c_d$. The first is the same as in misdetection, whereas the second is the classical case of
a local robustness query, which has been extensively explored~\cite{WaZhXuLiJaHsKo21,ZhWeChHsDa18}. The conjunction of both conditions is verified; if one of them does not hold, the network is not robust to the misclassification attack. 

Combining the two conditions by conjunction might lead to inefficiency, 
for example when the first condition takes a long time to verify while 
the second can be refuted much faster. To address this, we introduce a 
\emph{tool-agnostic alternation scheme}, given in~\cref{alg:alternation_for_misclassification}, 
which alternates between verifying the IoU predicate ($f^{\text{IoU}}$) 
and the class predicate ($f^{\text{mix}}$). 

\begin{algorithm}[ht!]
\caption{Tool-agnostic alternation for misclassification (with caching)}
\label{alg:alternation_for_misclassification}
\textbf{Input:} object detector $f$, initial timeout $T>0$
\begin{algorithmic}[1]
\State $boundsDet \gets \varnothing$,\; $boundsCla \gets \varnothing$
\Repeat
  \State $(\texttt{boundsDet},~res_{det}) \gets \text{verify-no-misdetection}\big(f^{\text{IoU}},~z \ge 0,~T,~\texttt{boundsDet}\big)$
  \State $(\texttt{boundsCla},~res_{clf}) \gets \text{verify-equal-prediction}\big(f^{\text{mix}},~c_g{=}c_d,~T,~\texttt{boundsCla}\big)$
  \If {$res_{det} = \textit{Safe}$ \& $res_{clf} = \textit{Safe}$}
    \State \Return \textit{Safe}
  \ElsIf {$res_{det} = \textit{Unsafe}$ \textbf{or} $res_{clf} = \textit{Unsafe}$}
    \State \Return \textit{Unsafe}
  \EndIf
  \State $T \gets 2\cdot T$
\Until{False}
\end{algorithmic}
\end{algorithm}

For clarity, we denote the original network by $f$, the IoU-augmented variant by $f^{\text{IoU}}$, and the combined IoU+class network by $f^{\text{mix}}$.
We maintain caches \texttt{boundsDet} and \texttt{boundsCla} (initialized empty and reused across iterations) to avoid redundant computation.
We use: \textbf{Safe} : property holds for all $x\in\mathcal{P}$; \textbf{Unsafe} : counterexample found; \textbf{Unknown} : timeout.
We encode $\mathrm{IoU}\ge\tau$ as $z=A_I-\tau A_U$ and verify $z\ge 0$.

The corresponding soundness and completeness guarantees are stated in ~\cref{prop:alternation}. 

\begin{proposition} \label{prop:alternation}
    \cref{alg:alternation_for_misclassification} is sound and complete.
\end{proposition}
\begin{proof}
    \cref{alg:alternation_for_misclassification} runs a loop with increased time in each iteration. Given an underlying sound and complete verification tool, the times $T_1,T_2$ it takes to the tool to solve each of the queries in lines 4,5 are finite (from its completeness). The loop in lines 3-12 eventually reach an iteration with timeout $T>max(T_1,T_2)$, because the initial (positive) timeout is multiplied by 2 in the end of each iteration (line 11), and in this iteration answers from the underlying verification tool for both queries are obtained. As a result, at least one of the conditions in lines 6 and 8 must hold, and \cref{alg:alternation_for_misclassification} terminates. Hence, it is complete. 
    
    When the algorithm returns an answer, it is \textit{Unsafe} if at least one of the answers returned by the underlying sound verification tool is \textit{Unsafe}, and it returns \textit{Safe} only if both answers are \textit{Safe}. Therefore, the final answer preserves the soundness of the underlying verification tool.
\end{proof}

\subsubsection{Overdetection.}
In general, this scenario closely resembles the previous one. By temporarily swapping the roles of the ground-truth ($G$) and predicted bboxes ($D$) in the definition of misclassification and applying the same verification process, we effectively check for over-detection.
In the single-object case, overdetection collapses to misclassification:
since $o_g=o_d=1$, both require $\mathrm{IoU} > \tau$ and $c_g=c_d$;
no additional cases exist.

\subsection{Extension to Multi-Object Detection}
\label{multi_OD}

In this section, we detail the extension of our encoding to support the verification of models that detect multiple objects in a single image. 
Assume that the ground truth objects are represented by the set of tuples \(\{G_1, G_2, \dots, G_{o_g}\}\), and the output of the object detection model is represented by \(o_d\) detected objects given as a set of tuples \(\{D_1, D_2, \ldots, D_{o_d}\}\).
For any pair of ground truth object $G_i$ and a detected object $D_j$, we can construct an encoding as described in~\cref{encoding}, terminating with the neuron $z_{ij}$. By our construction, $z_{ij}\geq0$  if and only if the IoU of $G_i$ and $D_j$ is greater than $\tau$.
Thus, we formally define the following verification problems:

\begin{enumerate}
    \item{\textbf{Misdetection}. To avoid misdetection, we need to ensure that all ground truth objects are detected by the model. The object $G_i$ is detected by some detected object $D_j$ if and only if for some $D_j$ we have that $z_{ij}\geq0$. This can be formulated as $\underset{j\in[o_d]}{\max}z_{ij} \geq 0$. Therefore, the misdetection property for all objects can be formulated as: 
    $\lnot Q_1 \coloneq o_g \leq o_d \wedge \underset{i\in[o_g]}{\bigwedge} \underset{j\in[o_d]}{\max}z_{ij} \geq 0$.
    }
    \item \textbf{Misclassification}. Unlike  the single-object case, when maximizing over $z_{ij}$ we cannot add a condition $c^g_i=c^d_j$, as it is not an optimization term that can be optimized with  the value of $z_{ij}$.   
    Instead, we denote $w_{ij}$, a fresh variable that equals zero if and only if $c^g_i=c^d_j$ and is negative otherwise. We first show how $w_{ij}$ is used, and then formalize it.
    We define:  \[s_{ij}=-ReLU(-z_{ij}) + w_{ij}\]   and note that $s_{ij}=0$ if $w_{ij} = 0$ and $z_{ij}\geq 0$, i.e., both classification and detection conditions  are 
    satisfied.  Otherwise, at least one component contributes a negative number to the sum and thus $s_{ij} < 0$. This can be computed within the DNN by extending it with a single  $ReLU$ neuron per $G_i,D_j$ pair, as in~\cite{ElGoKa20,ElCoKa22,CoElBaKa23,IsZoBaKa23}.
    This construction is illustrated in~\cref{Figure_multiod}.
    Finally, we encode the misclassification property as: 
    $\lnot Q_2 \coloneq o_g \leq o_d \wedge \underset{i\in[o_g]}{\bigwedge}\underset{j\in[o_d]}{\max}s_{ij} \geq 0$.
    We now formalize $w_{ij}$. Recall that a class is determined by the \emph{argmax} of its logit vector, thus $c^g_i=c^d_j$ if and only if the corresponding vectors share the same argmax. Denote $\bar{\alpha}, \bar{\beta}$ the vectors associated with $c^g_i,c^d_j$, respectively. For each index $k$, define $ a_k \coloneq \underset{l}{max}(\alpha_l - \alpha_k), b_k \coloneq \underset{l}{max}(\beta_l - \beta_k)$, and observe that  $a_k + b_k = 0$ if and only if $k$ is the argmax of both $\bar{\alpha},\bar{\beta}$, and is positive otherwise. 
    Lastly, we define $w_{ij} \coloneq \underset{k}{max} (-a_k - b_k) $ that equals zero if some $k$ is the shared argmax, and is negative otherwise. This formalization can be encoded within the network, by adding a \emph{max} neuron per class per $G_i,D_j$. It is omitted from~\cref{Figure_multiod} due to space limitations.

    \begin{figure}[h!]
    \centering
    \vspace{-0.4cm}\includegraphics[width=0.6\linewidth,keepaspectratio,clip,trim={0 0.3cm 0 0.5cm}]{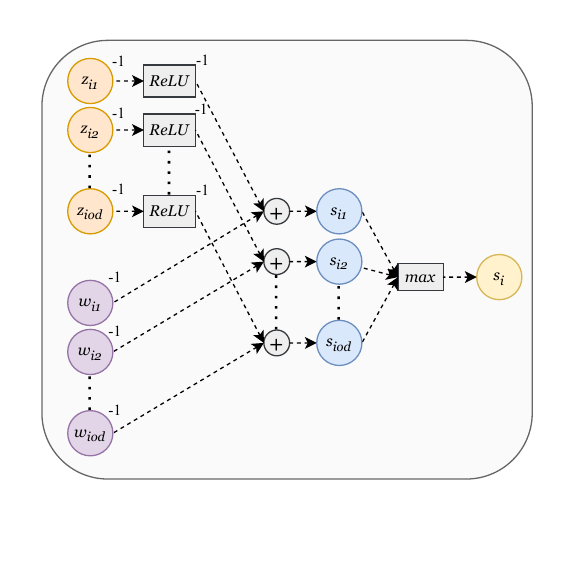}
    \vspace{-0.8cm}
    \caption{Encoding a misclassification property for the $i^{th}$ object in multi-object detection.}
    \label{Figure_multiod}
    \vspace{-0.4cm}
    \end{figure}

    \item{\textbf{Overdetection}. To avoid overdetection, each predicted object should detect some ground truth object. The object $D_j$ is detecting some ground truth object $G_i$ if and only if for some $G_i$ we have that $z_{ij}\geq0$. This can be formulated as $\underset{i\in[o_g]}{\max}z_{ij} \geq 0$. Therefore, the overdetection property for all objects can be formulated as:
    $\lnot Q_3 \coloneq o_d \leq o_g \wedge \underset{j\in[o_d]}{\bigwedge} \underset{i\in[o_g]}{\max}z_{ij} \geq 0$, or as  $o_d \leq o_g \wedge \underset{j\in[o_d]}{\bigwedge} \underset{i\in[o_g]}{\max}s_{ij} \geq 0$ to include classification.
    }
\end{enumerate}

\paragraph{Scope and limitations.}
We target pre-NMS verification and local robustness regions $\{x:\lVert x-x_0\rVert_p\le\epsilon\}$, with $p\in\{1,2,\infty\}$. Our encoding adds a lightweight head that realizes the predicate IoU$>\tau$ as $z=A_I-\tau A_U>0$ and combines it with class constraints; backends (e.g., Alpha-Beta-CROWN) then answer standard Safe/Unsafe/Unknown queries. We intentionally abstract away post-processing (e.g., NMS~\cite{GiDoDaMa13}) and large production detectors to isolate core verification issues. 
This scoped step provides complete (as well as sound) guarantees and actionable counterexamples.

\section{Evaluation}\label{sec:evaluation}

To evaluate the effectiveness of our approach, we run
 AlphaBeta-CROWN on a macOS (Apple M3, 16GB RAM, 8-core GPU). 
 We conduct experiments on datasets and architectures used in a state-of-the-art method~\cite{CoDuBoGaPaPuGa24}.

We run experiments on four datasets. \emph{LARD (Large Aviation Recognition Dataset)} provides $100{,}000+$ training and $10{,}000+$ test images of high-resolution aerial scenes for runway detection during aircraft approach/landing~\cite{DuCaFeGaMuPaSa23}; the large variability in ground-truth box sizes makes stability verification challenging. \emph{MNIST-OD} contains $60{,}000$ training and $10{,}000$ test images of size $90\!\times\!90$, each with a single MNIST digit (10 classes) randomly placed on a black background~\cite{De12}. \emph{GTSRB} includes $39{,}209$ training and $12{,}630$ test images of size $64\!\times\!64$ over 43 traffic-sign classes, with realistic variations in illumination, viewpoint, and distortion~\cite{StScSaIg11,HoStSaScIg13}. Finally, \emph{MNIST-2} is a smaller digit-detection set with $5{,}500$ training and $550$ test images (10 classes) at $64\!\times\!64$ resolution.

The trained object detectors
(architecture proposed by~\cite{CoDuBoGaPaPuGa24})
include 2--3
convolutional layers followed by 2--3 linear layers; as well as two
heads, one for detection and one for classification. We use CrossEntropyLoss for classification head and a SmoothL1 (Huber) loss for bbox regression, optimized with Adam ($lr=0.001$) for 10 epochs. No \IoU{} threshold was used during training; the \IoU{} threshold $\theta$ is used solely at verification/evaluation time to define correctness of detection. The details of
the models appear in~\cref{tab:trained_models}. We believe these models are comparable to the YOLO benchmark in \cite{vnncomp2025}, which uses a much smaller backbone and defines specifications on outputs before postprocessing.
\begin{table}[htbp]
\vspace{-0.8cm}
\centering
\caption{Model Overview}
\label{tab:trained_models}
\resizebox{0.8\columnwidth}{!}{%
\begin{tabular}{c|c|c|c|c|c|c}
    \toprule
    \multirow{2}{*}{\textbf{Network}} & \multirow{2}{*}{\textbf{\#Layers}} & \multicolumn{3}{c|}{\textbf{\#Parameters}} & \multicolumn{2}{c}{\textbf{Performance}} \\ \cmidrule{3-7}
    & {} & {\textbf{\small{Body}}} & {\textbf{\small{Classification head}}} & {\textbf{\small{Regression head}}} &{ \textbf{\small{Class Accuracy}}} & {\textbf{\small{Mean \IoU{}}}} \\ \midrule\midrule
    {LARD} & {5} & {16,887,104} & 516 & {-}    & {-}    & {0.52} \\ 
    {MNIST-OD} & {6} & {1,985,200}  & {2,570}    & {1,028}  & {98.4}    & {0.864} \\ 
    {GTSRB}    & {6} & {1,985,200}  & {2,570}    & {11,051} & {83.2}    & {0.732}  \\
    {MNIST-2}    & {3} & {1,082,760}  & {-}  & {1024} & {-}    & {0.551}  \\
\bottomrule\bottomrule
\end{tabular}
}
\end{table}

To dispatch the verification queries, we use 
Alpha-Beta-CROWN~\cite{WaZhXuLiJaHsKo21}, a state-of-the-art 
verification tool~\cite{MuBrBaLiJo23,BrBaLiJo23}, although other 
verifiers could in principle serve as backends. 
\cref{alg:alternation_for_misclassification} provides a 
tool-agnostic alternation scheme; when instantiated with 
Alpha-Beta-CROWN, it yields the refined version shown in~\cref{alg:alt-abc}. This instantiation exploits the three main 
phases of Alpha-Beta-CROWN: (i) adversarial attacks to quickly refute 
the property; (ii) incomplete verification to efficiently prove 
robustness when possible; and (iii) complete verification with 
branch-and-bound guided by the bounds from previous stages. 
By interleaving these phases for both IoU and class constraints, 
~\cref{alg:alt-abc} prioritizes fast refutations and proofs before 
resorting to the full alternation scheme.

\begin{algorithm}[ht!]
  \caption{Instantiation of ~\cref{alg:alternation_for_misclassification} using Alpha-Beta-CROWN}
  \label{alg:alt-abc}
  \textbf{Input:} Object detector $f$, initial timeout $T$
  \begin{algorithmic}[1]
    \State start $\gets$ get\_time()
    \State $res_{clf} \gets attack(f^{\text{mix}}, c_g = c_d)$
    \State $res_{det} \gets attack(f^{\text{IoU}}, z>0)$
    \If {$res_{det} = \textit{Unsafe}$ \textbf{or} $res_{clf} = \textit{Unsafe}$}
      \State \Return \textit{Unsafe}
    \EndIf
    \State $res_{clf} \gets$ Safe-incomplete($f^{\text{mix}}, c_g = c_d$)
    \State $res_{det} \gets$ Safe-incomplete($f^{\text{IoU}}, z>0$)
    \If {$res_{det} = \textit{Safe}$ \textbf{and} $res_{clf} = \textit{Safe}$}
      \State \Return \textit{Safe}
    \EndIf
    \State \Return ~\cref{alg:alternation_for_misclassification}($f$, $T - (get\_time()-start)$)
  \end{algorithmic}
\end{algorithm}

\begin{proposition}  \label{prop:alt-abc}
  ~\cref{alg:alt-abc} is sound and complete.
\end{proposition}
\begin{proof}
    ~\cref{alg:alt-abc} is sound and complete, as it inherits these 
  properties from the underlying attack, incomplete, and complete 
  stages of Alpha-Beta-CROWN together with   ~\cref{alg:alternation_for_misclassification}.
\end{proof}

\mysubsection{Results.}
We applied our method to assess each model's robustness with respect
to misdetection  and misclassification (which is equivalent to overdetection for
single-object). We conducted our experiments on a wide range of 17
perturbation sizes for each image among the first 100 samples in
each dataset, which sums up to 1700 total experiments for each
dataset.

\begin{figure*}[ht!]
    \centering
    \includegraphics[width=0.9\linewidth, keepaspectratio]{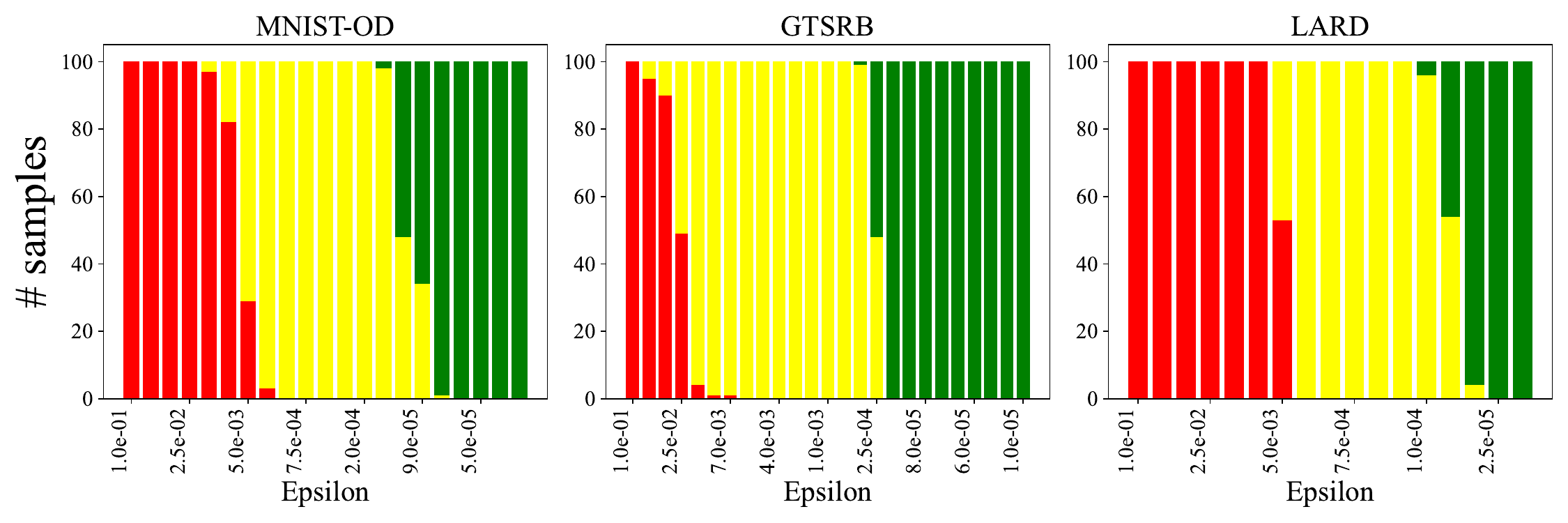}
    \includegraphics[width=0.9\linewidth, keepaspectratio]{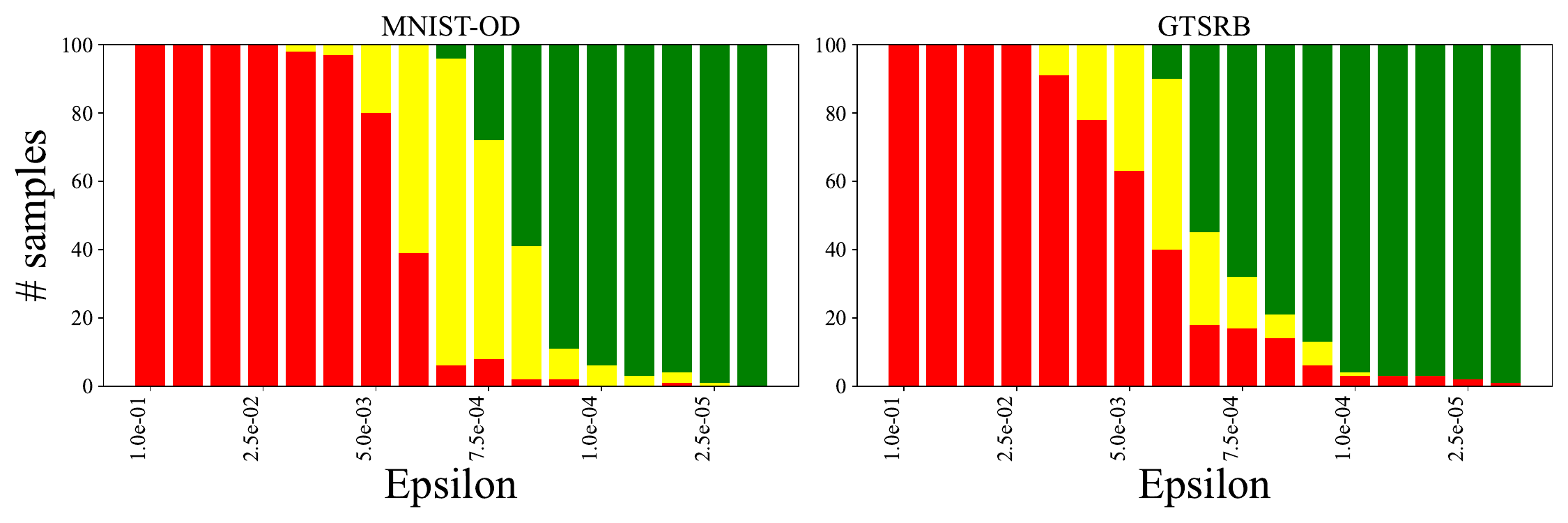}
    \caption{Misdetection (top) and misclassification (bottom) results per epsilon.}
    \label{fig:combined-results}
    \vspace{-0.4cm}
\end{figure*}

~\cref{fig:combined-results} shows the results for misdetection
robustness (top), and for misclassification robustness (bottom). Green, yellow and red colors
represent \textit{Safe}, \textit{Unknown} and \textit{Unsafe}
results, respectively. As epsilon (the maximal allowed perturbation) decreases, more robustness properties can be verified.

As a measure of runtime, we focused on the average runtime per sample. We
tested the 5 smallest and 5 largest epsilon values in
\cref{fig:combined-results} (bottom) in order to conduct a separate analysis for \textit{Unsafe} and \textit{Safe} instances. We ran the experiments with a five-minute timeout and successfully solved 88\% of the instances. The average runtime per solved sample for \textit{Unsafe} instances was $0.0052$ seconds, while the average for \textit{Safe} instances was $0.011$ seconds. These results demonstrate that both refuting and certifying the model’s robustness to misdetection and misclassification are highly efficient, with fast execution times across \textit{Unsafe} and \textit{Safe} instances.

\mysubsection{Comparison to IBP-IoU.}
We compare the performance of our method to that introduced in
\cite{CoDuBoGaPaPuGa24}, which employs interval bound propagation (IBP) 
for IoU computation in several configurations: Vanilla\_IoU,
Optimal\_IoU, CROWN-IBP + Vanilla\_IoU, and CROWN-IBP + Optimal\_IoU.\footnote{%
Note that \cite{CoDuBoGaPaPuGa24} also includes a version of CROWN 
without IBP, which uses CROWN to bound the output range without leveraging IBP's 
interval propagation. However, this setting was not supported for the 
$\epsilon$ parameters we experimented with.} 
Here is a brief explanation for each configuration:

\begin{enumerate}
  \item \textbf{Vanilla\_IoU:} Bounds the predicted bbox parameters 
        using IBP, and naively (soundly) bounds the IoU based on these intervals. 
  \item \textbf{Optimal\_IoU:} 
  Tightens IBP-derived bounds; it
        derives formulas for the exact minimum/maximum possible IoU over the interval range of the predicted box coordinates, determining the worst-case IoU with the ground-truth box.
  \item \textbf{CROWN-IBP + Vanilla\_IoU:} Combines CROWN-IBP 
        \cite{ZhHoChSvRoBoDuCh20}, which provides enhanced bounds for 
        the bbox variables, with the Vanilla\_IoU method. 
  \item \textbf{CROWN-IBP + Optimal\_IoU:} Integrates CROWN-IBP’s 
        bounding with the refined Optimal\_IoU method, affording the most 
        accurate IoU bounds among IBP-based variants.
\end{enumerate}

\begin{figure}[ht!]
\centering
 \vspace{-0.6cm}
 \includegraphics[width=0.8\linewidth, keepaspectratio]{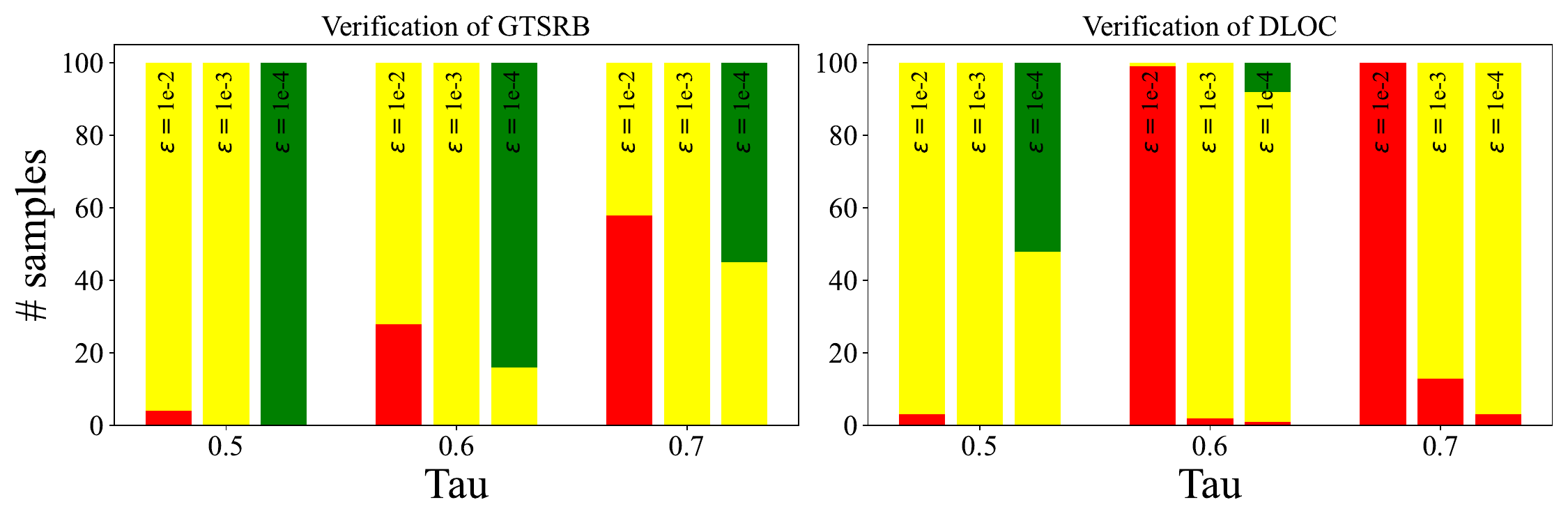}
 \vspace{-0.2cm}
 \caption{Effect of $\tau$. The number of \textit{Unsafe} results increases with $\tau$.}
 \label{diff_tau}
 \vspace{-0.6cm}
\end{figure}

\paragraph{Experimental setup.} 
All experiments are run with a \emph{timeout of 5 minutes per query}. 
We report the \emph{average runtime per solved sample}, separately for 
\textit{Safe} and \textit{Unsafe} cases. 
Unsafe instances are typically solved faster, as adversarial examples are 
often found in the initial attack phase. 
Unless otherwise stated, we use $p=\infty$. 
The $\epsilon$ values of local robustness queries were adopted from 
\cite{CoDuBoGaPaPuGa24}: 
LARD uses $\{4,6,8\}\times 10^{-4}$, and MNIST-OD uses 
$\{2,4,6\}\times 10^{-4}$. 
For~\cref{diff_tau}, we additionally evaluate 
$\tau \in \{0.5,0.6,0.7\}$ and 
$\epsilon \in \{10^{-2},10^{-3},10^{-4}\}$.

\paragraph{Precision.} 
Our framework inherits the floating-point precision of the underlying verifier 
(Alpha-Beta-CROWN, FP32 by default). 
To avoid edge cases, we verify inequalities of the form 
$z \ge \delta$ with a small $\delta$ (e.g., $10^{-8}$). 
This margin does not change conclusions but prevents numerical ties.

\paragraph{Results.} 
As presented in~\cref{tab:comparison_results}, our method stands out as 
the only approach that successfully identifies \textit{Unsafe} instances 
across both datasets. In total, our method solves 423 instances for MNIST-OD 
and 434 for LARD --- about 70\% additional instances compared to the runner-up 
(240 and 263, respectively).

\begin{table}[htbp]
    \centering
    \caption{Comparison of verification methods on \textit{MNIST-OD} and \textit{LARD} datasets, showing the number of \textit{Safe}, \textit{Unsafe}, and \textit{Unknown} results, along with average runtime.}
    \label{tab:comparison_results}
    \resizebox{0.9\columnwidth}{!}{%
    \begin{tabular}{c|c|c|c|c|c|c|c|c}
        \toprule
        \multirow{2}{*}{Method} & \multicolumn{4}{c|}{MNIST-OD} & \multicolumn{4}{c}{LARD} \\
        \cline{2-9}
        & Safe & Unsafe & Unknown & \shortstack{Average\\Time (sec)} & Safe & Unsafe & Unknown & \shortstack{Average\\Time (sec)} \\

        \midrule\midrule
        Vanilla\_IoU & 102 & 0 & 578 & 0.00010 & 61 & 0 & 659 & 0.00012 \\
        Optim\_IoU & 182 & 0 & 498 & 0.01143 & 90 & 0 & 630 & 0.01502 \\
        CROWN-IBP+Vanilla\_IoU & 207 & 0 & 473 & 0.00009 & 235 & 0 & 485 & 0.00009 \\
        CROWN-IBP+Optim\_IoU & 240 & 0 & 440 & 0.01099 & 263 & 0 & 457 & 0.01106 \\
        Ours & 183 & 240 & 257 & 2.73072 & 135 & 259 & 286 & 1.60556 \\
        \bottomrule\bottomrule
    \end{tabular}%
    }
\end{table}

In terms of runtime, our method is slower, with an average of 2.73s on MNIST-OD 
and 1.61s on LARD, compared to fractions of a second for IBP-IoU variants. 
However, this increase in computation time results in more comprehensive 
verification, particularly in finding unsafe scenarios, which are critical 
for robustness analysis.

In conclusion, while our method sacrifices some runtime efficiency (partly because Alpha-Beta-CROWN starts with an adversarial attack and only then propagates bounds), it offers 
a significant advantage by returning both \textit{Safe} and \textit{Unsafe} 
results and greatly increasing the number of solved queries. This suggests a 
portfolio approach, where faster but incomplete methods are run first, and 
our method is applied when they fail, providing a more reliable verification 
tool for neural networks.

\begin{figure}[h!]
\centering
\vspace{-0.2cm}
 \includegraphics[width=0.65\columnwidth, keepaspectratio]{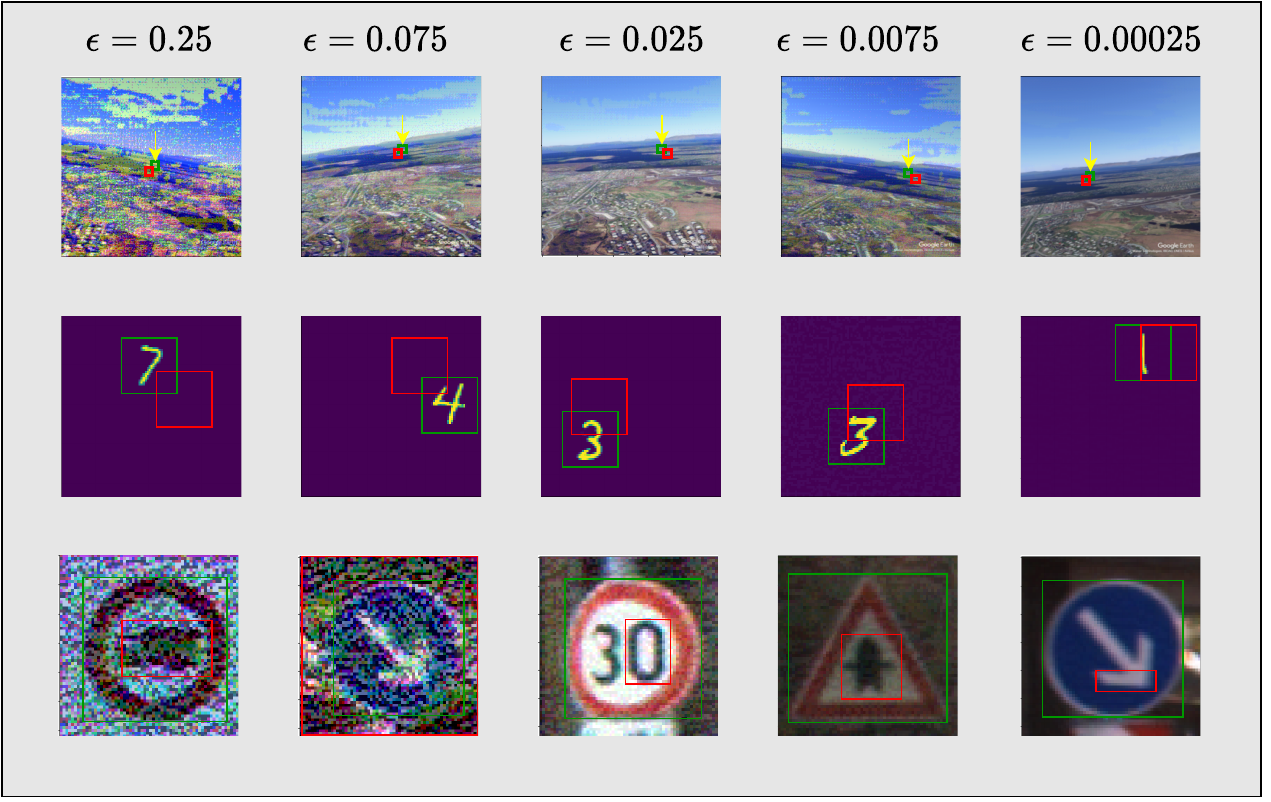}
 \caption{Misdetection examples on all datasets. Ground truth (green bboxes), with misdetected objects (red bboxes) under adversarial attacks of varying noise levels.}
 \label{Figure_misdet}
\end{figure}

\mysubsection{Qualitative Results.}
We provide further qualitative results in~\cref{Figure_misdet}. It illustrates multiple instances where our method identifies violation in misdetection robustness across different models, including high-resolution images. This occurs when the detected bboxes are either contained within, enclosing, or partially overlapping with the ground truth bbox. For instance, in the image located in the bottom-right corner, the signpost is not correctly detected. Instead of identifying the ground truth bbox (shown in green), only a partial region is detected (highlighted in red).

\mysubsection{Tau Effect.}
As discussed earlier in this paper, all initial experiments were conducted using an IoU matching threshold value of $\tau = 0.5$, which is a common assumption in the context of object detection tasks. In this subsection, we delve deeper into the effect of varying the value of $\tau$ on the formal verification process. Specifically, we explore how increasing $\tau$ influences the outcomes of the verification process under different attack conditions.

\cref{diff_tau} illustrates the results of experiments conducted on two distinct datasets, namely GTSRB and MNIST-OD, under a misdetection attack scenario. For this analysis, we evaluated three different values of $\tau$ in the set of $[0.5, 0.6, 0.7]$, and for each of these values, we tested three corresponding values of $\epsilon$: $\epsilon \in \{10^{-2}, 10^{-3}, 10^{-4} \}$. As demonstrated in the figure, an increase in the value of $\tau$ leads to a significant increase in the number of counterexamples found, while simultaneously reducing the number of successfully verified samples.

For instance, when using the GTSRB dataset and setting $\epsilon = 10^{-2}$, the number of counterexamples increased substantially, rising from 4 to 28, and then to 58, as $\tau$ was increased from 0.5 to 0.6 and then to 0.7. Similarly, when $\epsilon$ was set to $10^{-4}$, the number of verified examples decreased from 100 to 84, and further dropped to 55 as $\tau$ increased. These results align with our expectations, as a higher $\tau$ introduces stricter verification criteria, which in turn makes it easier to uncover potential vulnerabilities within the model. At the same time, it becomes increasingly difficult to prove the model's robustness under these more stringent conditions.

These findings highlight the critical role that the parameter $\tau$ plays in the verification process, particularly in terms of balancing the trade-off between finding vulnerabilities and demonstrating robustness. By adjusting $\tau$, we can effectively control the rigor of the verification process, making it a key factor in ensuring the safety and reliability of object detection models.

\mysubsection{Multi-Object Detection.}
Lastly, we extend our evaluation to multi-object detection case, described in~\cref{multi_OD}.
As a proof of concept, we trained a simple DNN to detect two objects (without classification) on the MNIST-2 dataset~\cite{MNIST2}, which consists of images containing two MNIST digits. More details about the model are given in~\cref{tab:trained_models}.
We then extended the trained network with the encoding described in~\cref{multi_OD} tailored to two-object detectors.

\begin{figure}[ht!]
\centering
\scalebox{0.65}{
\vspace{-0.8cm}
 \includegraphics[width=0.35\linewidth, keepaspectratio,valign=c]{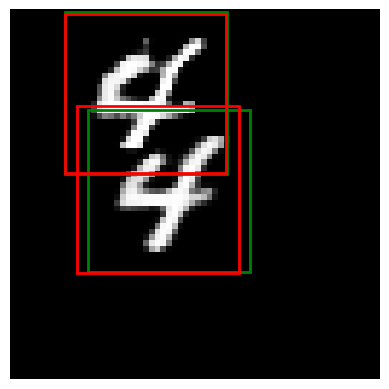}
 \includegraphics[width=0.65\linewidth, keepaspectratio,valign=c]{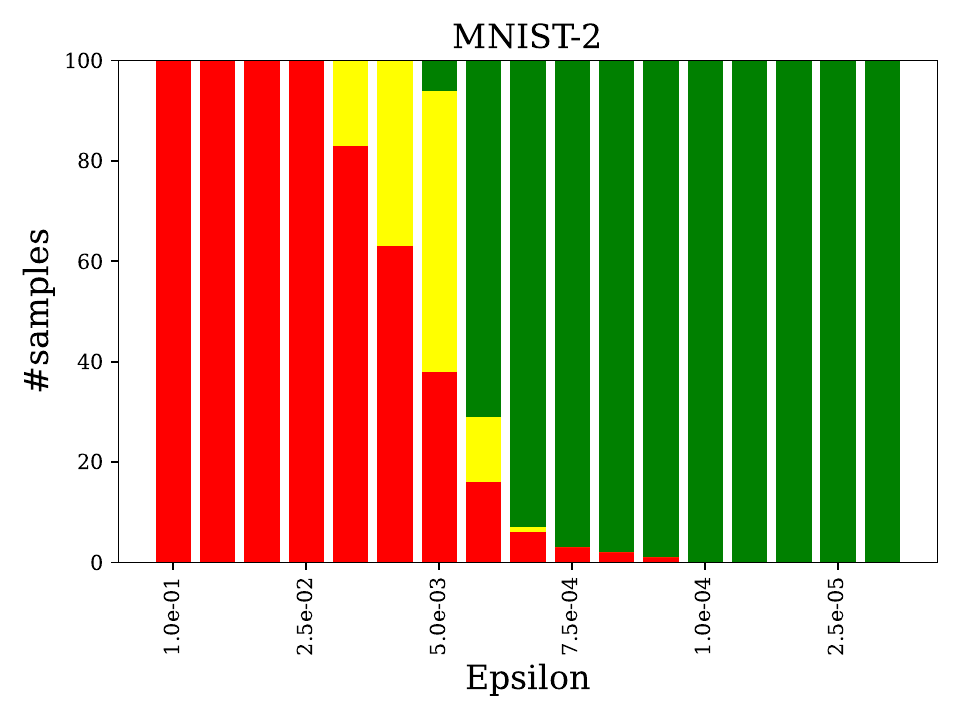}
 }
\vspace{-0.4cm}
 \caption{(Left) Example of a correct two-object detection in an MNIST-2 sample. (Right) Misdetection results per epsilon, for a single object in a two-objects image.}
\label{mod_results}
\vspace{-0.8cm}
\end{figure}

Next, we applied our method to assess the model’s robustness to misdetection of an arbitrary single digit by either of the two detection heads, using the same experimental policy as in the single-object detection case.
\cref{mod_results} shows the results for misdetection robustness, as well as an example of an MNIST-2 image.
In the image sample, the red boxes represent the ground truth detection, and green boxed represent the model's prediction. In the graph, green, yellow and red colors represent Safe, Unknown and Unsafe results, respectively.

\section{Conclusion}
\label{sec:Summary}

This work tackles the significant challenge of applying formal verification techniques to ensure the safety and reliability of computer vision models, with a specific emphasis on expanding these methods beyond traditional image classification tasks to encompass the more complex domain of object detection. In particular, it introduces a comprehensive framework that formalizes a range of potential attack types, including misdetection, misclassification, and overdetection, by framing these issues as formal verification problems within the context of classification models. 

Furthermore, the study develops a detailed methodology to apply state-of-the-art verification tools in order to verify the correctness and robustness of object detection models. The proposed approach is thoroughly evaluated across multiple models and datasets, setting an important benchmark that can guide and inform future research in this area. This evaluation demonstrates the feasibility of incorporating formal verification into the workflow of computer vision systems, offering a foundation for future advancements.

The contributions of this work significantly extend the existing framework for formal verification of deep neural networks (DNNs), as evidenced by the implementation and case study examples presented throughout the study. While additional research is required to refine and expand these methods, the findings presented here establish a 
foundation for the integration of formal verification techniques into the field of computer vision. This integration holds the potential to enhance the safety, security, and reliability of these technologies, ultimately contributing to their more responsible and trustworthy deployment in real-world applications.

\section*{Acknowledgements}
The work of Elboher, Isac and Katz was partially funded by the European Union (ERC, VeriDeL, 101112713). Views and opinions expressed are however those of the author(s) only and do not necessarily reflect those of the European Union or the European Research Council Executive Agency. Neither the European Union nor the granting authority can be held responsible for them. The work of Elboher, Isac and Katz was additionally supported by a grant from the Israeli Science Foundation (grant number 558/24).
The work of Raviv, Aluf-Medina and Kugler was supported by the Horizon 2020 research and innovation programme for the Bio4Comp project under grant agreement number 732482, by the Israel Science Foundation (Grant No. 190/19) and by the Data Science Institute at Bar-Ilan University (seed grant). We thank and acknowledge the contributions of Yael Leibovich Weiss, Omer Cohen and Roy Assa who were involved in earlier versions of this work.

\bibliographystyle{unsrt}
\bibliography{NFM2026}

\end{document}